%% file: m9687.tex
\documentclass[camera-ready]{ecai} 

\usepackage{latexsym}
\usepackage{amssymb}
\usepackage{amsmath}
\usepackage{amsthm}
\usepackage{booktabs}
\usepackage{enumitem}
\usepackage{graphicx}
\usepackage{color}

\usepackage{algorithm}
\usepackage{algpseudocode}
\usepackage{algorithmicx}
\usepackage[switch]{lineno}
\usepackage{mathtools}
\usepackage{amssymb}
\usepackage[hidelinks]{hyperref}
\usepackage[utf8]{inputenc}
\usepackage[small,skip=5pt]{caption}
\usepackage{xcolor,colortbl}
\usepackage{multirow}
\usepackage{float}
\usepackage{cleveref}

\newcommand{\BibTeX}{B\kern-.05em{\sc i\kern-.025em b}\kern-.08em\TeX}

\begin{document}


\begin{frontmatter}


\paperid{9687} 


\title{Enabling Weak Client Participation via On-Device Knowledge Distillation in Heterogeneous Federated Learning}



\author[A]{\fnms{Jihyun}~\snm{Lim}}
\author[A]{\fnms{Junhyuk}~\snm{Jo}}
\author[B]{\fnms{Tuo}~\snm{Zhang}} 
\author[A]{\fnms{Sunwoo}~\snm{Lee}\thanks{Corresponding Author. Email: sunwool@inha.ac.kr}}
\address[A]{Inha University, South Korea}
\address[B]{University of Southern California, United States}

\begin{abstract}
    Online Knowledge Distillation (KD) is recently highlighted to train large models in Federated Learning (FL) environments.
    Many existing studies adopt the logit ensemble method to perform KD on the server side.
    However, they often assume that unlabeled data collected at the edge is centralized on the server.
    Moreover, the logit ensemble method personalizes local models, which can degrade the quality of soft targets, especially when data is highly non-IID.
    To address these critical limitations, we propose a novel on-device KD-based heterogeneous FL method.
    Our approach leverages a small auxiliary model to learn from labeled local data.
    Subsequently, a subset of clients with strong system resources transfers knowledge to a large model through on-device KD using their unlabeled data.
    Our extensive experiments demonstrate that our on-device KD-based heterogeneous FL method effectively utilizes the system resources of all edge devices as well as the unlabeled data, resulting in higher accuracy compared to SOTA KD-based FL methods.
\end{abstract}
\end{frontmatter}

\input{1_intro}

\input{2_background}

\input{3_design}
\input{4_eval}

\input{5_conclusion}

\section*{Acknowledgments}
This work was supported by INHA UNIVERSITY Research Grant.


\bibliography{mybibfile}

\end{document}

%% file: 1_intro.tex
\section{Introduction}

Federated Learning (FL)~\cite{mcmahan2017communication} is a practical learning method to utilize the data collected on-the-fly at the edge.
With recent advances in the system efficiency of edge devices, various large-scale on-device learning scenarios are being considered, such as multi-task transfer learning~\cite{chen2019edge}, sharpness-aware minimization~\cite{fan2024locally}, and even Large Language Model (LLM) fine-tuning~\cite{zhu2023pockengine}.
These studies generally assume homogeneous devices, where all participants can contribute equally to model training.
However, FL environments often consist of heterogeneous edge devices, such as mobile phones and IoT devices, which continuously gather new, unlabeled local data~\cite{tashakori2023semipfl,chen2023distributed} that could be used for training.
Therefore, to exploit the ever-increasing big data collected at the edge, it is imperative to develop a new FL method that can fully utilize heterogeneous system resources.



Several recent studies tackle the system heterogeneity issue in FL by employing partial model training scheme~\cite{diao2020heterofl,horvath2021fjord,lee2023layer,lee2023partial,lee2024embracing,zhang2023timelyfl}.
However, they enforce clients with weak resources to train only a subset of the target model's layers or channels, significantly harming the freedom of model design.
More recent works utilize Knowledge Distillation (KD)~\cite{hinton2015distilling}.
They commonly use a logit ensemble method instead of directly aggregating the local models to reduce the KD's communication cost in FL environments~\cite{jeong2018communication,itahara2021distillation}.
In our empirical study, however, we find that the logit ensemble fails to provide sufficiently accurate soft targets when the data distribution is highly non-IID.
Although a large amount of public data is available, KD with such inaccurate soft targets negatively impacts the efficiency of knowledge transfer.
Some others utilize unlabeled extra data to train the target model on the server side~\cite{lin2020ensemble,cho2022heterogeneous}.
This approach assumes that unlabeled data generated on edge devices is aggregated on the server and utilized for centralized training, raising privacy concerns.

In this paper, we propose a novel heterogeneous FL method that utilizes unlabeled data at the edge to overcome the limitations of logit ensemble methods and server-side model training approaches.
Specifically, we introduce a KD-based FL method that jointly utilizes additional system resources on strong devices and unlabeled local data by combining auxiliary model training and on-device KD.
The core principle behind our method is that a small auxiliary model first learns global knowledge through standard model aggregation and then helps the training of the large target model by utilizing system resources and unlabeled local data on strong devices.
When transferring global knowledge from the auxiliary model to the target model, only a subset of clients with sufficiently strong system resources participate in the KD process.
This heterogeneous approach allows all clients to effectively join the training, exploiting their own resources and local data.

Compared to logit ensemble-based KD, our method has two strong benefits as follows.
First, our approach ensures accurate soft targets for the KD process.
By aggregating the small auxiliary models across clients, the local knowledge is merged into a single point, thereby significantly improving the quality of the soft targets compared to the logit ensemble.
Second, our approach does not require the server to have sufficiently strong system capability nor the unlabeled data to be centralized on the server in advance.
By offloading the KD workload to the strong devices, the model can learn from the local data regardless of whether it is labeled or not.
Moreover, as more clients join the training, our approach allows the model to learn from more unlabeled data, likely improving the model accuracy.

We evaluate the performance of our proposed method across various machine learning benchmarks: CIFAR-10~\cite{krizhevsky2009learning}, 
 CIFAR-100 (fine-tuning), FEMNIST~\cite{caldas2018leaf}, IMDB reviews~\cite{zm1y-b270-20} , and Google Speech Commands~\cite{warden2018speech}.
Our extensive experiments demonstrate that the proposed method outperforms SOTA KD-based FL methods under highly non-IID conditions.
We also theoretically analyze the generalization bound for the proposed method.
The contributions of our work are summarized below.
\begin{itemize}
    \item We explain and empirically prove the fundamental limitations of the logit ensemble-based Knowledge Distillation especially when data is strongly non-IID.
    \item Our study demonstrates how to harmonize online Knowledge Distillation with FL to better utilize unlabeled data collected at the edge on-the-fly.
    \item We discuss how to exploit heterogeneous system resources in realistic FL environments via selectively offloading the distillation workload to the edge.
    \item Our study presents an extensive comparative analysis, demonstrating that the proposed method achieves superior soft target quality, resulting in higher accuracy.
\end{itemize}

%% file: 2_background.tex
\section {Related Works}

\subsection {Heterogeneous Federated Learning}
In realistic FL environments, some devices may not efficiently train large models or even cannot put them in their memory space.
In this case, there are two options: employ a small model giving up the learning capability or keep using the large model giving up the private data owned by the weak devices.
One promising solution to this system heterogeneity issue is partial model training method~\cite{diao2020heterofl,horvath2021fjord,lee2023layer,lee2023partial,lee2024embracing,zhang2023timelyfl,alam2022fedrolex}.
While these methods alleviate the workload for some clients, they do not consider independent model architectures across clients.

Another promising approach is heterogeneous model approximation method.
Some recent works allow clients to reduce the model size according to their system capability~\cite{yao2021fedhm,liu2023hetefed,chan2021fedhe,niu2022federated}.
However, the low-rank model approximation causes a non-negligible extra computational cost.
Additionally, the local models cannot be directly aggregated because each one may be approximated to a different rank.

\subsection {Knowledge Distillation in Federated Learning}
Table~\ref{tab:sota} summarizes the recently proposed KD-based FL methods.
First, there are two homogeneous FL methods: FedAvg and MOON~\cite{li2021model}.
Neither of them considers the newly collected unlabeled data or utilizes KD.
\\
\textbf{Logit Ensemble} --  KD has been widely used in recent FL studies with a goal of either improving the model accuracy or training a large model.
Federated KD proposed in ~\cite{jeong2018communication} directly exchanges the soft targets across the server and the clients instead of exchanging the model parameters.
DS-FL proposed in~\cite{itahara2021distillation} exchanges only the soft targets under a strong assumption that the unlabeled data is available for all individual clients.
FedMD~\cite{li2019fedmd} transfers knowledge by leveraging the fine-tuning mechanism.
These methods commonly aggregate the local models' output logits at the server and broadcast them to the clients.
This approach significantly reduces the communication cost; however, the accuracy drop is significant.
\\
\textbf{Logit Ensemble with Server-Side Distillation} -- 
FedDF~\cite{lin2020ensemble} trains the target model at the server-side using the logit ensemble method.
This approach is under an assumption that a large amount of unlabeled data is already available at the server side.
It also assumes the server has sufficient system resources to directly run KD.
FedFTG~\cite{zhang2022fine} also uses the server-side KD together with a feature generator.
This study does not consider the system heterogeneity nor the additional unlabeled data.
\\
\textbf{Logit Ensemble, Server-Side Distillation, and Heterogeneous Models} -- FedGKT~\cite{he2020group} allows the server to have its own classifier layers and train them using KD.
This approach achieves promising model accuracy, however, it does not take into account locally generated unlabeled data.
FedGEMS~\cite{cheng2021fedgems} also runs KD to train a large model at the server-side.
Like FedMD, this method also assumes the public data to be labeled.
More recently, Fed-ET~\cite{cho2022heterogeneous} has been proposed, which utilizes model transfer as well as weighted ensemble at the server-side.
However, this approach also performs KD on the server side, and the server-side model cannot directly learn from the clients' private data.

\begin{table}[t]
\footnotesize
\centering
\begin{tabular}{lcccc} \toprule
\multirow{2}{*}{Method} & Server & \multirow{2}{*}{Exchange} & Additional Data & \multirow{2}{*}{KD} \\ 
& Model & & / Labeled & \\ \toprule
FedAvg & =client & model & No / N/A & N/A \\
MOON & =client & model & No / N/A & N/A \\
FD & N/A & logits & No / N/A & device \\
DS-FL & N/A & logits & Yes / No & device \\
FedMD & N/A & logits & Yes / Yes  & device \\
FedDF & =client & model & Yes / No & server \\
FedGEMS & $>$ client & logits & Yes / Yes & both \\
FedGKT & $>$ client & model+logits & No / N/A & both \\
Fed-ET & $>$ client & model+logits & Yes / No & server \\
\textbf{Ours} & N/A & model & \textbf{Yes / No} & \textbf{device} \\ \bottomrule
\end{tabular}
\caption{
    The configurations of SOTA FL methods.
}
\label{tab:sota}
\end{table}

%% file: 3_design.tex
\section {Method} \label{sec:method}
We begin with an ablation study to demonstrate how logit ensemble can harm KD performance in general FL environments.
Based on this analysis, we develop our novel on-device KD-based FL method, which maximizes KD performance by fully utilizing the system resources of heterogeneous devices.
Finally, we analyze the expected generalization bound for the proposed method and discuss how unlabeled local data impacts its performance.

\subsection {Limitations of Logit Ensemble in Federated Learning Environments} \label{sec:ablation}
Many existing works employ the logit ensemble method to implement communication-efficient online KD in FL environments~\cite{jeong2018communication,itahara2021distillation,lin2020ensemble,cho2022heterogeneous}.
However, our empirical study finds that it can significantly hurt the model accuracy.
Our research is strongly motivated by this ablation study result.

We compare the accuracy of logit ensemble and that of the conventional model aggregation.
We use CIFAR-10 and Google Speech Command, which are computer vision and audio recognition benchmarks.
First, the dataset is partitioned into two distinct subsets: one for training and the other for evaluating model accuracy.
We further divide the training data into 100 small subsets using a label-based Dirichlet distribution ($\alpha=0.1$) and assign one small subset to each client.
Next, we conduct logit ensemble-based FL and model aggregation-based FL, and compare the accuracies they achieve.
Note that the logit ensemble uses the full model, while the model aggregation uses a smaller model, with each layer containing $25\%$ of the channels.
This setting penalizes the model aggregation scheme in terms of the model's learning capability, and we expect to see the adverse impact of the logit ensemble more clearly under this setting.

\begin{figure}[t]
\centering
\includegraphics[width=\columnwidth]{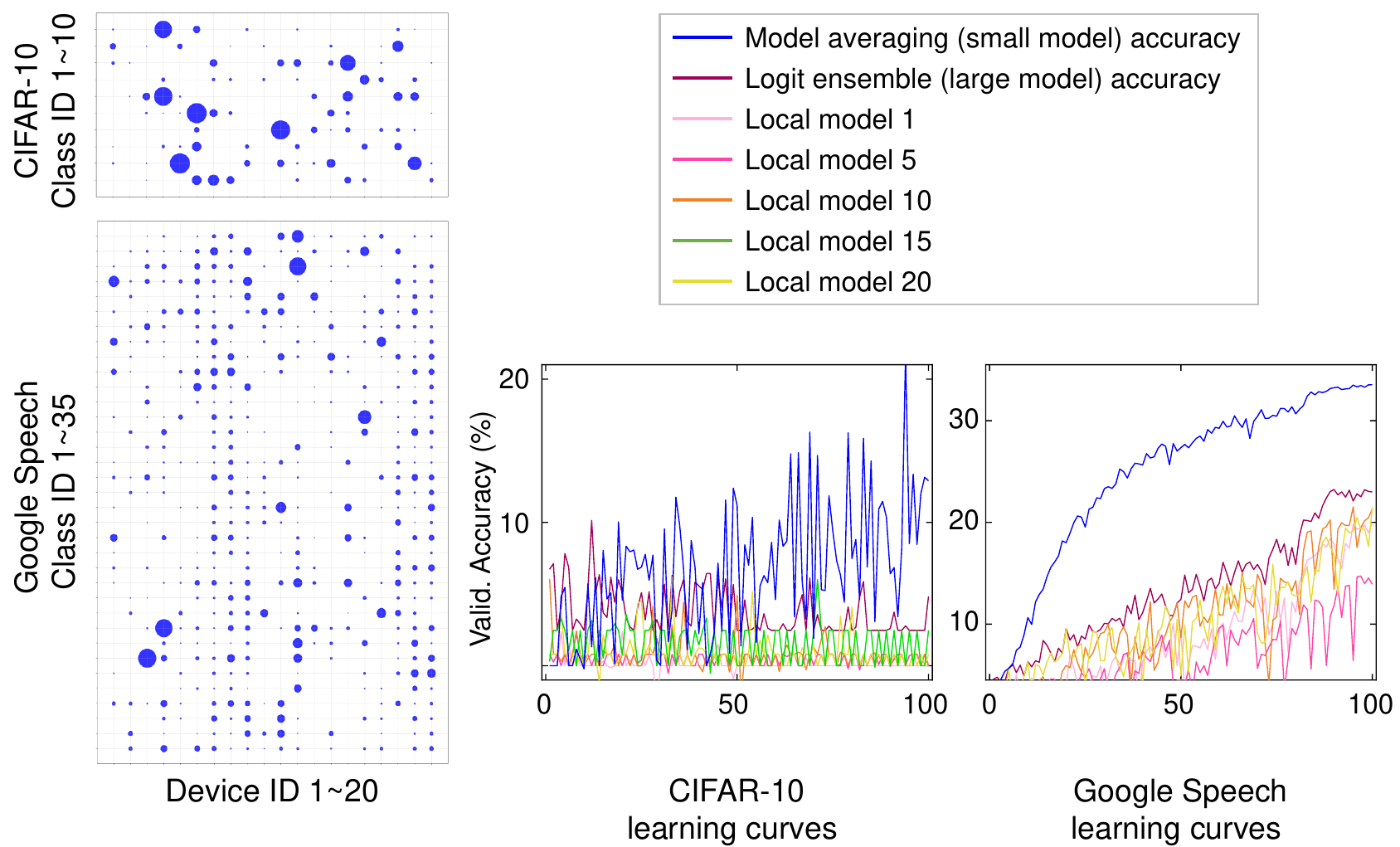}
\caption{
    Non-IID distributions of CIFAR-10 (left-top) and Google Speech (left-bottom). Label-based Dirichlet distributions ($\alpha=0.1$) are used to get 100 local distributions. Due to the limited space, we only show 20 of them. The accuracy of a small model trained via FL provides remarkably more accurate logits than the logit ensemble.
}
\vspace{2em} 
\label{fig:distribution}
\end{figure}

In Figure~\ref{fig:distribution}, the left bubble chart shows the class-wise data distribution per client, while the right side of the figure shows the learning curve comparisons.
The bubble charts illustrate how non-IID the datasets are.
Most clients have data from only a few classes, indicating highly non-IID conditions.
Each learning curve chart shows the accuracies of five local models as well as the global model's accuracy.
The purple curve corresponds to the global model trained with the logit ensemble method, while the blue curve corresponds to the global model trained with conventional model aggregation.
The gap between the global curve and the local curves quantifies the impact of each FL method.

When data is highly non-IID, the logit ensemble accuracy of the full model (purple curves) is considerably lower than the model aggregation accuracy of the small ($25\%$) model (blue curves).
\textit{The logit ensemble forces all local models to be personalized without synchronizing model parameters.}
Such personalized models lead to a significant degradation in the soft target quality in terms of generalization.
Some previous works assume that labeled public data is available at the edge as a way to address this issue; however, having a large amount of labeled public data is unrealistic~\cite{li2019fedmd,cheng2021fedgems}.
In addition, new unlabeled data is most likely generated at the edge, and centralizing them to the server may pose privacy concerns.
For these reasons, the logit ensemble is not an appropriate option for heterogeneous FL.
Motivated by this analysis, we propose a novel on-device KD-based FL method that addresses the low-quality soft target issue.

\begin{figure*}[t]
\centering
\includegraphics[width=1.6\columnwidth]{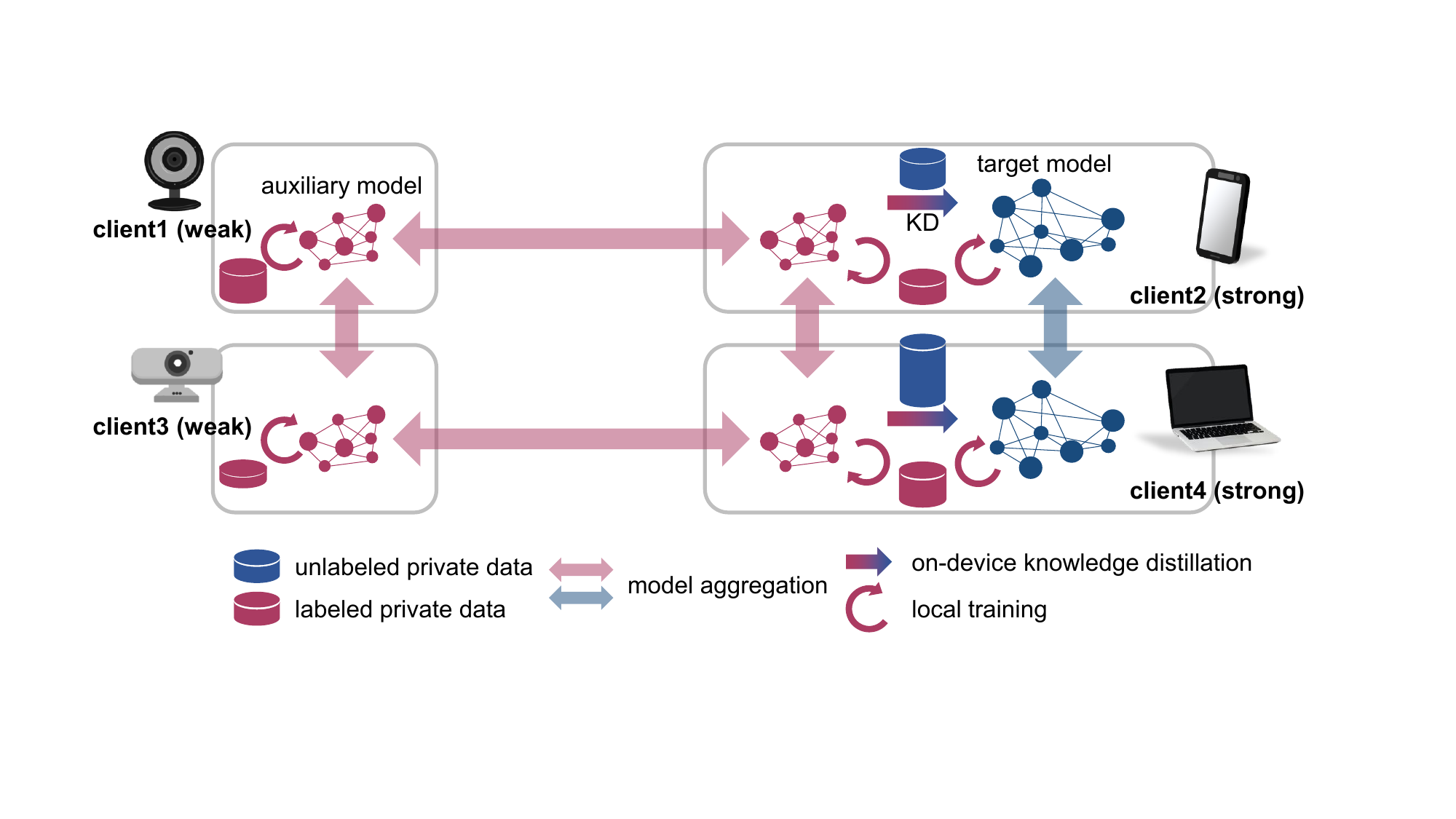}
\caption{
    The schematic illustration of the proposed heterogeneous FL method. For simplicity, the server is omitted in this schematic. An auxiliary (small) model is trained by all the devices using their private data. Then, the global knowledge is transferred to the target (large) model through on-device KD. This distributed knowledge transfer approach enables to not only fully utilize heterogeneous edge devices' system resources but also effectively use the unlabeled private data for training.
}
\vspace{1em} 
\label{fig:schematic}
\end{figure*}

\subsection {On-Device Federated Transfer Learning} \label{sec:FTL}
We first define two types of models: $\textbf{w}_l$ and $\textbf{w}_s$.
\begin{itemize}
    \item{$\textbf{w}_l$: The large target model.}
    \item{$\textbf{w}_s$: A smaller auxiliary model.}
\end{itemize}
We also define two relative categories of clients: \textit{strong} and \textit{weak} clients.
A client is considered \textit{strong} if it has sufficient system resources to train the large target model $w_l$.
On the other hand, a client is considered \textit{weak} if it cannot independently train the large target model $w_l$.
This limitation may be due to various system constraints, such as insufficient memory or too slow computational resources.
We do not specify the exact reason in order to keep the definition of this type general.

Now, we focus on federated optimization, which trains the target model $\textbf{w}_l$.
Considering the natural data collection at the edge, we assume that each edge device has a certain amount of unlabeled prviate data.
This is a valid problem formulation because it is more reasonable for new data to be generated at the edge rather than the parameter server.
Our method employs KD with these unlabeled local data to improve the target model accuracy.
We take advantage of an auxiliary model $\textbf{w}_s$, which fits within the memory capacity of all individual devices.

Given the two models, our method repeats the following two steps until a user-defined termination condition is satisfied (e.g., target accuracy or time budget).
Figure~\ref{fig:schematic} provides a schematic illustration of these two steps.
\\
\textbf{Step 1: Auxiliary Model Training Round} -- First, a random subset of clients is activated in each round.
The activated devices receive $\textbf{w}_s$ from the server and run $\tau$ local updates using their own labeled data.
The resulting locally trained models, $\textbf{w}_s^i$ for $i \in A$, are aggregated at the server using simple uniform averaging, where $A$ is the set of active clients.
This global FL round with the small auxiliary model enables the collection of local knowledge from all clients, regardless of their system capacity.
The loss function in step 1 is as follows. 
\begin{align}
    \mathcal{L}_s(\textbf{w}_s) = \frac{1}{n} \sum_{i=1}^{n} \textrm{CE}(\textbf{w}_s, x), \forall x \in \hat{\mathcal{D}}^i,
\end{align}
where $\textbf{w}_s$ is the auxiliary model, $n$ is the total number of clients, $\textrm{CE}(\cdot, \cdot)$ is the cross entropy function, and $\hat{\mathcal{D}}^i$ is the local data owned by client $i$.
\\
\textbf{Step 2: Target Model Training Round} -- Once the auxiliary models are aggregated at the server, the strong clients train the target model $\textbf{w}_l$ using the auxiliary model.
First, the strong clients receive both models, $\textbf{w}_s$ and $\textbf{w}_l$ from the server.
Then, they locally update the target model $\textbf{w}_l$ for $\tau$ steps using their own labeled data.
After the local updates, they run additional $\tau$ local KD steps using the unlabeled private data.
The auxiliary model participates in the KD process as a teacher.
Finally, the local target models $\textbf{w}_l^j, \forall j \in S$ are aggregated at the server, where $S$ is the set of strong clients.

In step 2, all individual strong clients train the target model $\textbf{w}_l^j$ using the following loss functions. 
\begin{align}
    \mathcal{L}_{l}(\textbf{w}_l) &= \frac{1}{s}\sum_{j=1}^{s} \textrm{CE}(\textbf{w}_l, x), \forall x \in \hat{\mathcal{D}}^j \nonumber \\
    & \hspace{0.5cm} + \frac{1}{s}\sum_{j=1}^{s} \lambda \cdot \textrm{KL}(\textbf{w}_s, \textbf{w}_l, x'), \forall x' \in \mathcal{D}^j,
\end{align}
where $\textbf{w}_l$ is the target model, $s$ is the number of strong clients, $\textrm{KL}(\cdot, \cdot, \cdot)$ is the Kullback–Leibler divergence function, and the $\lambda$ is a hyper-parameter that determines how strongly the $\textrm{KL}(\cdot, \cdot, \cdot)$ affects the total loss.
Note that the first term on the right-hand side uses the local labeled data $x \in \hat{\mathcal{D}}^j$ while the second term uses the local unlabeled data $x' \in \mathcal{D}^j$.

\subsection {Discussion on Design Choices of On-Device Knowledge Distillation}
\textbf{The Role of Auxiliary Model} -- 
The core principle of our heterogeneous FL method is to use a small auxiliary model trained with a conventional model aggregation scheme.
This design choice ensures local data privacy, as the auxiliary model is trained locally, and only its parameters are aggregated.
In contrast, logit ensemble assumes that clients can access the same unlabeled data (also known as public data), enabling logits from different local models to be computed using the same input data.
Our method does not depend on this strong assumption and allows each client to fully utilize their local data, whether labeled or unlabeled.

In addition, the logit ensemble allows local models to remain personalized, as model parameters are not directly averaged across clients.
If data is highly non-IID -- for example, if each client has local data limited to one or two classes -- these personalized models are likely to suffer from poor generalization performance.
Our ablation study results are shown in Figure~\ref{fig:distribution}.
Thus, we can conclude that aggregating local logits alone is insufficient to provide accurate soft targets for KD.
\\
\textbf{Comparison to Server-side Knowledge Distillation} -- 
Our on-device KD method has three advantages over conventional server-side distillation.
First, it enables the target model to utilize both labeled and unlabeled data on the clients.
This is a critical benefit in terms of scalability, as the target model can learn from more unlabeled data as the number of strong clients grows.
In other words, our method is better suited for on-device learning, where new data is continuously generated at the edge.
Second, our method takes advantage of system heterogeneity, allowing clients with stronger system resources to run more training and distillation steps than others.
This approach naturally maximizes system utilization at the edge.
Finally, our method enables each client to locally utilize new data collected at runtime without sharing it with others.
Some logit ensemble-based KD methods~\cite{lin2020ensemble,cho2022heterogeneous,cheng2021fedgems} assume that \lq{}public data\rq{} are accessible at both clients and server.
Such an assumption is overly strong and unrealistic, as it may pose privacy issues.
\\
\textbf{Communication Cost Analysis} -- We compare the communication cost of our method with that of representative logit ensemble-based KD methods.
Table~\ref{tab:comm} presents the results of the comparative study conducted on CIFAR-10 (ResNet20).
FedDF employs server-side KD, whereas DS-FL adopts client-side KD.
To ensure a fair comparison, we align the total number of rounds, regardless of whether they involve training or KD.

For the homogeneous FL methods (FedAvg and FedDF), we consider two cases: one case involves training the small model $w_s$ (with $25\%$ of the channels at every layer) on all clients, while the other case trains the large model $w_l$ only on a subset of clients with sufficiently strong system resources. First, FedAvg (small) has a lower communication cost than FedAvg (large), but suffers from significantly lower accuracy.
Interestingly, the accuracy of FedDF is lower than that of FedAvg, regardless of the model size.
This result indicates that the soft targets generated by the logit ensemble are of insufficient quality to make meaningful progress in KD.
FedDF can leverage KD without incurring extra communication cost.
However, it runs KD on the server-side with inaccurate soft targets, which could ultimately harm model accuracy.
DS-FL, the client-side KD-based FL method, adopts the logit ensemble and entirely avoids the model aggregation cost.
Thus, the communication cost is only the number of classes $C$, but the model accuracy is significantly reduced.

The communication cost of our method is slightly higher than that of FedAvg(large) due to the extra communication for the auxiliary model training.
However, the accuracy improves remarkably, even surpassing that of FedAvg (large).
This analysis demonstrates that our method strikes a practical trade-off between communication cost and model accuracy.
\\
\textbf{Computational Cost Analysis} --
We also compare the computational cost of our method to that of the same logit ensemble-based KD methods discussed in Table~\ref{tab:comm}.
Since FedDF and FedAvg are homogeneous FL methods that consider only a single type of model, a fair comparison of their computational costs with that of our method is not possible.
Therefore, we focus on the training cost of $w_l$ only.
Our method has the same computational cost as FedAvg (large) and FedDF (large), while achieving significantly higher accuracy.
DS-FL is a heterogeneous FL method and its computational cost is directly comparable to our method.
However, its accuracy is significantly lower than all the other methods due to the ineffective knowledge transfer through logit ensemble.
This computational cost comparison suggests that our auxiliary model improves the target model's accuracy without significantly increasing the total computational cost.

\begin{table}[t]
\footnotesize
\centering
\begin{tabular}{lcccc} \toprule
Method & Comm. Cost & Message Size & Acc ($\%$) \\ \midrule
FedAvg(small) & $|w_s|$ & 19K & $57.47 \pm 0.8\%$ \\
FedAvg(large) & $|w_l|$ & 274K & $59.21 \pm 1.5\%$ \\ 
FedDF(small) & $|w_s|$ & 19K & $ 41.62\pm 0.1\%$ \\
FedDF(large) & $|w_l|$ & 274K & $58.94\pm 0.1\%$ \\
DS-FL & $C$ & $10$ & $25.95 \pm 0.1\%$  \\
$\textbf{On-Device KD}$ & $|w_s|$ + $|w_l|$ & 19K + 274K  & 
$\textbf{62.43} \pm 0.1\%$  \\ \bottomrule
\end{tabular}
\caption{
    The communication cost comparison across different FL methods. $C$ is the number of classes. $w_l$ is ResNet20 while $w_s$ is a modified one which has only $25\%$ channels at every layer.
}
\vspace{1em} 
\label{tab:comm}
\end{table}

\subsection {Discussion on Impact of On-Device Knowledge Distillation on Generalization Performance}

Here, we theoretically analyze the benefit of our design choices in terms of the target model's generalization bound.
Our analysis builds upon that of \cite{cho2022heterogeneous}, extending it to account for the combination of labeled and unlabeled local data.
\\
\textbf{Problem Definition} -- We first define hypothesis $h: \mathcal{X} \rightarrow \mathcal{Y}$, with input space $\mathbf{x} \in \mathcal{X}$ and label space $y \in \mathcal{Y}$, and the hypothesis space $h \in \mathcal{H}$.
The general loss function $l(h(\mathbf{x}), y)$ quantifies the classification error of the given hypothesis.
The true global data distribution is defined as $\mathcal{D}$.
Then, the true local data distribution from each client's perspective is defined as $\mathcal{D}_i, i \in [n]$.
The data distribution of the limited number of local observations is defined as $\mathcal{\hat{D}}_i$.
We also define the expected loss over an arbitrary data distribution $\mathcal{D}'$ as $\mathcal{L}_{\mathcal{D'}}(h) = \mathbb{E}_{\mathbf{x},y\sim \mathcal{D}'}\left[ l(h(\mathbf{x}), y) \right], \forall h \in \mathcal{H}$.
Finally, we assume that $\mathcal{L}(h)$ is convex.
In these settings, the generalization bound of the target model trained by our proposed method satisfies the following condition.

Now, to make our analysis self-contained, we state the two lemmas proposed in the previous work, which will be used to analyze the generalization of our proposed method.
\\
\textbf{Lemma 1 (Domain Adaptation~\cite{ben2010theory})}
\textit{With two true distributions $\mathcal{D}_{1}$ and $\mathcal{D}_{2}$, $\forall p \in (0,1)$ and hypothesis $\forall h \in \mathcal{H}$, with probability at least $1 - p$ over the choice of samples, it satisfies}
\begin{align}
    \mathcal{L}_{\mathcal{D}_1}(h) \leq \mathcal{L}_{\mathcal{D}_2}(h) + \frac{1}{2}d(\mathcal{D}_1, \mathcal{D}_2) + v_1,
\end{align}
\textit{where $d(\mathcal{D}_1, \mathcal{D}_2)$ quantifies the distribution discrepancy between the two data distributions, $\mathcal{D}_1$ and $\mathcal{D}_2$, $v_1$ is the minimal loss such that} $v_1 = \textrm{inf}_h \mathcal{L}_{\mathcal{D}_1}(h) + \mathcal{L}_{\mathcal{D}_2}(h)$.
\\
\textbf{Lemma 2 (Generalization with limited training samples)}~\cite{cho2022heterogeneous}.
\textit{For all individual client $i$, with probability at least 1 - p over the choice of samples, there exists:}
\begin{align}
    \mathcal{L}_{\mathcal{D}_i}(h_{\mathcal{\hat{D}}_i}) \leq \mathcal{L}_{\mathcal{\hat{D}}_i}(h_{\mathcal{\hat{D}}_i}) + \sqrt{\frac{log 2/p}{2|\mathcal{\hat{D}}_i|}},
\end{align}
\textit{where $|\mathcal{\hat{D}}_i|$ is the size of client $i$'s local dataset.}

The first lemma measures the generalization bound for one data distribution based on the maximum discrepancy between the given two distributions.
The second lemma elaborates the bound taking the limited training samples into account based on Hoeffding's inequality.
Our proposed method trains the target model $\mathbf{x}_l$ using $s < n$ strong clients only.
Thus, we will analyze the generalization bound with respect only to the strong clients' labeled data using the above two lemmas.
Then, we will revise this bound concerning the on-device knowledge distillation with unlabeled local data.
\\
\textbf{Proposition 1} \textit{Consider $n$ clients in total and $s$ of them are strong clients. Each client has its own data distribution $\mathcal{D}_i, \forall i \in [n]$. We define a combination of the labeled local data and the unlabeled local data as $\Tilde{\mathcal{D}}_i$. For an arbitrary data distribution $\mathcal{D}'$, $h_{\mathcal{D}'}$ is the model best-trained on $\mathcal{D}'$. Then, with probability of $1 - p$, our proposed on-device knowledge distillation-based FL guarantees the following generalization bound for the target model.}
\begin{align}
    \mathcal{L}_{\mathcal{D}} \left( \frac{1}{s} \sum_{i=1}^{s} h_{\mathcal{\Tilde{D}}_i} \right) &\leq \frac{1}{s}\sum_{i=1}^{s} \left( \mathcal{L}_{\Tilde{\mathcal{D}}_i}\left(h_{\mathcal{\Tilde{D}}_i} \right) 
 + \sqrt{\frac{\textrm{log} 2/ p}{2 | \Tilde{\mathcal{D}_i} |}} \right) \nonumber \\
   & \hspace{0.5cm} + \frac{1}{2s}\sum_{i=1}^{s} d(\mathcal{D}_i, \mathcal{D}) + \frac{1}{s}\sum_{i=1}^{s} v_i. \nonumber
\end{align}

\begin{proof}
First, let us bound the loss over the true global data distribution $\mathcal{L}_D(\frac{1}{s}\sum_{i=1}^{s} h_{\mathcal{\hat{D}}_i})$ based on lemma 1.
Note that the hypothesis is the model averaged only across $s$ strong clients.
\begin{align}
    \mathcal{L}_{\mathcal{D}} \left( \frac{1}{s} \sum_{i=1}^{s} h_{\mathcal{\hat{D}}_i} \right) &\leq \frac{1}{s} \sum_{i=1}^{s} \mathcal{L}_{\mathcal{D}} \left( h_{\mathcal{\hat{D}}_i} \right) \nonumber \\
    & \leq \frac{1}{s} \sum_{i=1}^{s} \left( \mathcal{L}_{\mathcal{D}_i} (h_{\mathcal{\hat{D}}_i}) + \frac{1}{2} d(\mathcal{D}_i, \mathcal{D}) + v_i \right). \label{bound1}
\end{align}

Then, (\ref{bound1}) can be further bounded using lemma 2 as follows.
\begin{align}
    \mathcal{L}_{\mathcal{D}} \left( \frac{1}{s} \sum_{i=1}^{s} h_{\mathcal{\hat{D}}_i} \right) &\leq \frac{1}{s}\sum_{i=1}^{s} \left( \mathcal{L}_{\hat{\mathcal{D}}_i}\left(h_{\mathcal{\hat{D}}_i} \right) 
 + \sqrt{\frac{\textrm{log} 2/ p}{2 | \hat{\mathcal{D}_i} |}} \right) \nonumber \\
   & \hspace{0.5cm} + \frac{1}{2s}\sum_{i=1}^{s} d(\mathcal{D}_i, \mathcal{D}) + \frac{1}{s}\sum_{i=1}^{s} v_i. \nonumber
\end{align}

Finally, the on-device knowledge distillation utilizes not only the strong clients' labeled data but also the unlabeled data.
Thus, we can simply replace the local dataset $\mathcal{\hat{D}}_i$ with the combination of the labeled and unlabeled data $\mathcal{\Tilde{D}}_i$.
\begin{align}
    \mathcal{L}_{\mathcal{D}} \left( \frac{1}{s} \sum_{i=1}^{s} h_{\mathcal{\Tilde{D}}_i} \right) &\leq \frac{1}{s}\sum_{i=1}^{s} \left( \mathcal{L}_{\Tilde{\mathcal{D}}_i}\left(h_{\mathcal{\Tilde{D}}_i} \right) 
 + \sqrt{\frac{\textrm{log} 2/ p}{2 | \Tilde{\mathcal{D}_i} |}} \right) \nonumber \\
   & \hspace{0.5cm} + \frac{1}{2s}\sum_{i=1}^{s} d(\mathcal{D}_i, \mathcal{D}) + \frac{1}{s}\sum_{i=1}^{s} v_i.\label{bound2}
\end{align}
\end{proof}
\textbf{Remark 1.} \textit{This result highlights the key advantage of our method: on-device KD reduces the second term on the right-hand side of (\ref{bound2}), leading to better generalization.
Note that $|\hat{\mathcal{D}}_i| \leq |\Tilde{\mathcal{D}}_i|$, where $\hat{\mathcal{D}}_i$ denotes the labeled data and $\Tilde{\mathcal{D}}_i$ denotes the union of labeled and unlabeled data.
Therefore, by expanding the data distribution, the proposed method can achieve a lower empirical loss.}

\textbf{Remark 2.} \textit{As each client collects more unlabeled data, the size of $\mathcal{\Tilde{D}}_i$ may increase and the bound decreases even further.
This is a critical benefit in realistic FL environments because the new unlabeled data is most likely collected at the edge rather than the server.
Thus, our method can be considered as a more scalable FL method especially when the strong clients can collect data on-the-fly.
Note that in the server-side KD, the $\mathcal{\hat{D}}_i$ can only be the unlabeled data which is still smaller than $\mathcal{\Tilde{D}}_i$.}

%% file: 4_eval.tex
\section {Experiments} \label{sec:eval}
\begin{table}[t]
\footnotesize
\centering
\begin{tabular}{lcc} \toprule
\multirow{2}{*}{Dataset} & Auxiliary Model (\textit{weak}) & \multirow{2}{*}{Target Model (\textit{strong})} \\
& (relative model size) &  \\ \midrule \midrule
\multirow{2}{*}{CIFAR-10} & ResNet20 & ResNet20 \\
& $\{25\%, 50\%, 75\%\}$ & $100\%$ \\ \midrule
\multirow{2}{*}{FEMNIST} & CNN & CNN \\
& $\{25\%, 50\%, 75\%\}$ & $100\%$ \\ \midrule
CIFAR-100 & ResNet50 & ViT-b16 \\
 (Fine-tuning) & $\sim 29\%$ & $100\%$ \\ \midrule
\multirow{2}{*}{IMDB} & Bi-LSTM & Bi-LSTM \\
& $25\%$ & $100\%$ \\ \midrule
\multirow{2}{*}{Google Speech} & CNN & CNN \\
& $\{25\%, 50\%, 75\%\}$ & $100\%$ \\ \bottomrule
\end{tabular}
\caption{
    The model configurations for all five benchmarks.
}\label{tab:clients}
\end{table}

\begin{table*}[t]
\footnotesize
\centering
\begin{tabular}{llcccccc} \toprule
Dataset & Batch (LR) & Local Steps & Epochs & \# of clients & Weak-only & Strong-only & Proposed \\ \midrule
CIFAR-10 & 32 (0.2) & 30 & 400 & 100 & $57.47\pm 0.8\%$ & $59.21 \pm 1.5\%$ & $\textbf{62.43} \pm 0.1\%$ \\ 
FEMNIST & 20 (0.02) & 30 & 200 & 100 & $64.27 \pm 0.1\%$ & $53.75 \pm 0.6\%$ & $\textbf{67.44} \pm 0.1 \%$ \\ 
CIFAR-100 (Fine-tuning) & 32 (0.04) & 60 & 40 & 100 & $40.56 \pm 0.1\%$ & $52.17 \pm 0.1\%$ & $\textbf{59.73} \pm 0.1\%$ \\ 
IMDB review & 10 (0.4) & 20 & 90 & 50 & $81.46 \pm 0.1\%$ & $59.63\pm 0.1\%$ & $\textbf{82.50} \pm 0.1\%$ \\
Google Speech & 32 (0.02) & 40 & 100 & 100 & $66.83 \pm 0.3\%$ & $68.83 \pm 0.1\%$ & $\textbf{71.95} \pm 0.5\%$ \\ \bottomrule
\end{tabular}
\caption{
    The FL performance comparison: weak model-only, strong model-only and the proposed method.
}\label{tab:base}
\end{table*}



\begin{figure*}[ht!]
\centering
\includegraphics[width=2\columnwidth]{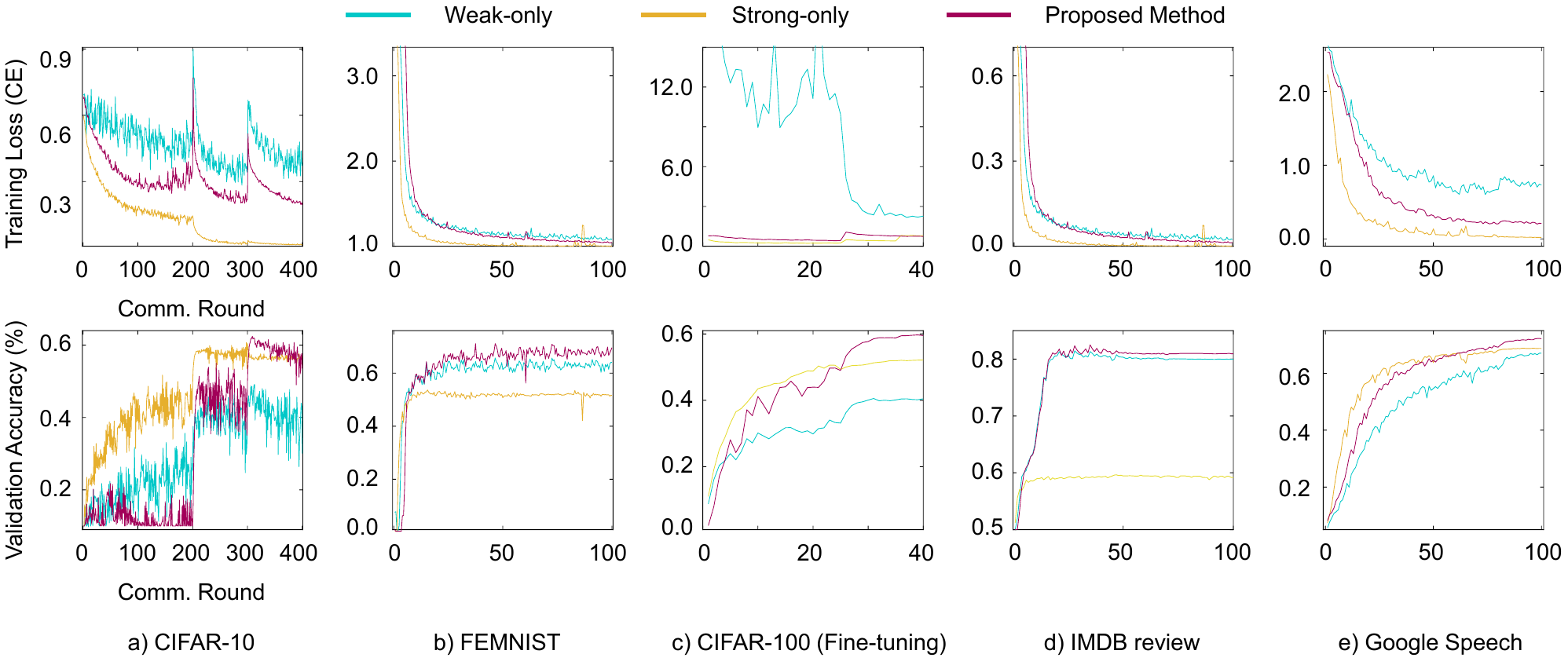}
\caption{
    The training loss (top) and validation accuracy (bottom) curves corresponding to Table~\ref{tab:base}.
}
\vspace{2em} 
\label{fig:curves}
\end{figure*}

\begin{table*}[t]
\footnotesize
\centering
\begin{tabular}{lccccccc} \toprule
\multirow{2}{*}{Dataset} & FedAvg & MOON & FedDF & DS-FL & FedMD & FedGEM & Proposed \\ 
 & (homo) & (homo) & (homo) & (hetero) & (hetero) & (hetero) & (hetero) \\ 
Public Data & N/A & N/A & Unlabeled & Unlabeled & Labeled & Labeled & \textbf{Unlabeled} \\ \midrule
CIFAR-10 & $57.47 \pm 0.8\%$ & $58.38 \pm 0.1\%$ & $41.62 \pm 0.1\%$ & $25.95 \pm 0.1\%$  &$61.13 \pm 0.1\%$  &$34.27 \pm 0.1\%$   & $\textbf{62.43} \pm 0.1\%$ \\ 
FEMNIST &$64.27 \pm 0.1\%$  & $64.80 \pm 0.1\%$ & $66.90 \pm 0.2\%$ & $56.71\pm 0.5\%$& $65.23\pm 0.1\%$ &$65.06 \pm 0.1\%$ & $\textbf{67.44} \pm 0.1\%$ \\ 
Google Speech & $66.05 \pm 0.7\%$ & N/A & $59.58 \pm 0.2\%$ & $19.14 \pm 0.1\% $& $65.90 \pm 0.1\%$ & $25.20 \pm 0.2\%$ & $\textbf{71.95} \pm 0.5\%$ \\ \bottomrule
\end{tabular}
\caption{
    The FL performance comparison across SOTA KD-based FL methods. $20\%$ of the total clients are \textit{strong} clients and the auxiliary model size is $25\%$ of the target model.
    Our method achieves the best accuracy without extra labeled data (public data). MOON is designed only for image data, and thus we do not apply it to Google Speech, the audio benchmark.
}
\label{tab:compare}
\vspace{2em} 
\end{table*}

We conduct performance evaluations in computer vision: CIFAR-10 (ResNet20~\cite{he2016deep}), CIFAR-100 (ViT~\cite{dosovitskiy2020image}), FEMNIST (CNN), natural language processing: IMDB (Bidirectional LSTM), and audio recognition: Google Speech Command (CNN) benchmarks.
We report accuracy averaged across at least 3 independent runs. For all the methods in our comparative study, the algorithm-specific hyper-parameters were highly tuned using appropriate grid searches.
See Appendix for the hyper-parameter settings in detail~\footnote{Appendix is available at https://www.arxiv.org/abs/2503.11151}.
\\
\textbf{Non-IID Dataset Settings} -- We use label-based Dirichlet distributions ($\alpha = 0.1$ which represents highly non-IID FL environments).
FEMNIST is naturally non-IID, and we use 100 clients' data as unlabeled local data.
Google Speech is also naturally non-IID, but each local dataset is quite large.
So, we use an artificial distribution to make it more strongly non-IID.
We activate random $20\%$ of the total clients at every round.
In all our experiments, half of the dataset is distributed to all clients as unlabeled local data.

\subsection {Performance Evaluation}
In all our experiments, we follow the definitions of \textit{strong} and \textit{weak} clients as discussed in Section~\ref{sec:FTL}.
As shown in Table~\ref{tab:clients}, we explore a variety of auxiliary model sizes for a comprehensive empirical study.

Table~\ref{tab:base} shows the performance comparison between two baselines and the proposed method, and Figure~\ref{fig:curves} shows the corresponding training loss and validation accuracy curves, respectively.
We consider two baselines: \textit{weak-only}, where the model size is reduced to fit within all clients' memory space and \textit{strong-only}, where the \textit{weak} clients are excluded and only the \textit{strong} clients participate in the large model training.
These two baselines are the most straightforward options for standard FL.
By comparing our method to these two baselines, we can evaluate how well the proposed on-device KD utilizes the \textit{weak} clients' local data to strengthen the large target model.
To ensure a fair comparison in terms of computational cost, the \textit{weak-only} and \textit{strong-only} methods perform $2\tau$ local steps, while our method runs $\tau$ local SGD steps and $\tau$ KD steps.

Table~\ref{tab:base} shows that the proposed method remarkably improves the target model accuracy.
The performance gap mainly comes from the knowledge transferred from the auxiliary model which learns from all clients' local data.
The same result can be observed across different applications, computer vision benchmarks and audio command detection benchmark.
It also works well with the transformer-based fine-tuning task.
Therefore, we can conclude that the proposed method effectively improves the large model's performance by utilizing many \textit{weak} clients' local data.

\subsection {Comparative Study}
We compare our method to three homogeneous FL methods: FedAvg, MOON, and FedDF, and heterogeneous FL methods: DS-FL, FedMD, and FedGEM~\cite{cheng2021fedgems}, as shown in Table~\ref{tab:compare}.
The homogeneous FL methods use a small model that has $25\%$ of channels at each layer.
The heterogeneous FL methods use the full model at $20\%$ of the clients and the small model at the rest of the clients.
These settings well represent the realistic FL environments consisting of edge devices with heterogeneous system resources.

We do not compare our method to FedGKT~\cite{he2020group} and Fed-ET~\cite{cho2022heterogeneous} since they are composed of several features independent of KD.
E.g., their representation transfer can be applied to our method without any conflicts.
We also do not present time vs. accuracy results because all the methods have a different computation to communication ratio per round.
Our study instead focuses on the achievable accuracy within a sufficiently large epoch budget.
\\
\textbf{Comparison to Homogeneous FL Methods} -- 
MOON achieves higher accuracy than FedAvg, demonstrating a practical use-case of contrastive loss.
FedDF achieves even higher accuracy than MOON thanks to its server-side KD with public data.
However, because these methods do not consider the system heterogeneity, they cannot train large models that do not fit into weak clients' memory space.
In contrast, our method can train large target models, achieving superior performance by exploiting the strong clients' system resources.
\\
\textbf{Comparison to Heterogeneous FL Methods} --
DS-FL shows much lower accuracy than others due to the personalized local models.
This result is well aligned with our ablation study shown in Figure~\ref{fig:distribution}, indicating that the logit ensemble dramatically harms the soft target quality, leading to a loss in accuracy.
FedMD achieves much higher accuracy than DS-FL.
However, the performance gain mainly comes from the labeled public data shared across all the clients.
In addition, FedMD is not applicable to environments where unlabeled data is distributed to the edge since the global logit should be calculated from the local logits obtained from the same data across all the clients.
FedGEM achieves inconsistent performance across benchmarks.
Due to the poor soft target quality caused by the logit ensemble, it severely loses the accuracy.
FedGEM also assumes the labeled public data like FedMD, which is not realistic.
Based on this comparative study, we can conclude that our method utilizes the unlabeled data more effectively than other methods.

\subsection {Ablation Study}
\textbf{Auxiliary Model Size} -- To analyze the impact of the auxiliary model size on the target model's performance, we conduct an ablation study as shown in Table~\ref{tab:size}.
First, we set the number of strong clients to $20\%$ of the total clients.
Then, we measure the accuracy of the auxiliary and target models using different auxiliary model sizes.
One clear trend observed across all settings is that the target model accuracy improves as the auxiliary model size increases.
This trend demonstrates that \textit{the auxiliary model more effectively transfers the global knowledge to the target model as its size grows}.

One intriguing observation is that, for CIFAR-10 and Google Speech, the auxiliary model achieves even higher accuracy than the target model when its size is $75\%$.
This is the case where the amount of data plays a more important role in achieving good accuracy than the model size difference.
That is, the increased learning capability from a larger target model does not significantly improve performance, as the sufficiently strong auxiliary model can already learn global knowledge effectively.
This result shows that the proposed method becomes more practical as the system becomes strongly heterogeneous.
\\
\textbf{Strong Client Ratio} -- We conduct another ablation study to analyze the impact of strong client ratio on FL performance.
Table~\ref{tab:ratio} shows the target model's accuracy with a varying number of strong clients.
The auxiliary model size is $25\%$ of the target model.
The unlabeled data size is fixed regardless of the number of strong clients.
In all the benchmarks, the target model performance is improved as the ratio of the strong clients increases.
This result demonstrates the benefit of having an auxiliary model: \textit{As the ratio of the strong clients increases, the target model can learn from more unlabeled data, which results in improving its validation accuracy}.
Therefore, our method can be considered as a promising heterogeneous FL method which scales up more effectively.

\begin{table}[t]
\footnotesize
\centering
\begin{tabular}{cccc} \toprule
Dataset & Ratio & Auxiliary Acc. & Target Model Acc.  \\ \midrule
\multirow{2}{*}{CIFAR-10} & $25\%$ & $49.11 \pm 0.1\%$ & $62.43 \pm 0.1\%$ \\ 
\multirow{2}{*}{ResNet20}& $50\%$ & $60.86\pm 1.5\%$& $64.62\pm 0.5\%$ \\
& $75\%$ & $66.65\pm 0.6\%$& $66.39 \pm 0.8\%$ \\ \midrule
\multirow{2}{*}{FEMNIST} & $25\%$ & $64.19 \pm 0.1\%$ & $67.44 \pm 0.1\%$ \\ 
\multirow{2}{*}{CNN}& $50\%$ & $65.89 \pm 0.4\%$ & $67.73 \pm 0.2\%$ \\
& $75\%$ & $66.67 \pm 0.3\%$ & $68.63 \pm 0.3\%$\\ \midrule
\multirow{2}{*}{Google Speech} & $25\%$ & $66.83 \pm 0.3\%$ & $71.95 \pm 0.5\%$ \\ 
\multirow{2}{*}{CNN} & $50\%$ & $74.54 \pm 0.1\%$ & $75.25 \pm 0.1\%$ \\
& $75\%$ & $76.77 \pm 0.1\%$ & $75.94 \pm 0.2\%$ \\ \bottomrule
\end{tabular}
\caption{
    The accuracy comparison across varying auxiliary model sizes. $20\%$ of the total clients are strong. 
}
\label{tab:size}
\vspace{1em} 
\end{table}

\begin{table}[t]
\footnotesize
\centering
\begin{tabular}{ccc} \toprule
Dataset & Strong / Total Clients & Target Model Acc.  \\ \midrule
\multirow{2}{*}{CIFAR-10} & 10 / 100 & $64.37 \pm 0.1\%$ \\ 
\multirow{2}{*}{ResNet20} & 20 / 100 & $68.29 \pm 0.7\%$ \\
& 40 / 100 & $70.20 \pm 0.8\%$ \\ \midrule
\multirow{2}{*}{FEMNIST} & 10 / 100 & $66.36 \pm 0.1\%$\\ 
\multirow{2}{*}{CNN} & 20 / 100 & $67.44 \pm 0.1\%$\\
& 40 / 100 & $67.78 \pm 0.1\%$\\ \midrule
\multirow{2}{*}{Google Speech} & 10 / 100 & $67.91 \pm 0.1\%$ \\ 
\multirow{2}{*}{CNN} & 20 / 100 & $71.95 \pm 0.5\%$ \\
& 40 / 100 & $74.72 \pm 0.4\%$ \\ \bottomrule
\end{tabular}
\caption{
    The accuracy comparison across varying strong client ratios. The auxiliary model size is $25\%$.
}
\label{tab:ratio}
\vspace{1em} 
\end{table}

%% file: 5_conclusion.tex
\section {Conclusion}
We proposed a novel on-device KD-based heterogeneous FL method.
Our method exploits the heterogeneous system resources at the edge via local KD with periodic model aggregations.
This approach enables the target model to learn from the local data regardless of whether they are labeled or not.
This is the key advantage as it will allow FL applications to leverage the continuously generated local data at the edge without needing to share it across clients.
Our study demonstrates that the proposed on-device KD method also effectively improves the knowledge transfer efficiency even under extremely non-IID environments.
More importantly, our study empirically proves that the large model can be well trained on the distributed non-IID data without relying on the server-side resources or the centralized public data.
We believe that harmonizing our method and the communication-efficient model aggregation methods is critical future work.